\documentclass[sigconf, 10pt, nonacm]{acmart}
\PassOptionsToPackage{table,xcdraw}{xcolor}
\usepackage{natbib}
\usepackage{tcolorbox}
\usepackage{fontawesome}
\usepackage{booktabs}
\usepackage{varwidth}
\usepackage{fvextra}
\usepackage{amsmath,amsfonts}
\usepackage{algorithmic}
\usepackage{graphicx}
\usepackage{xcolor}
\usepackage{textcomp}
\usepackage{float}
\usepackage{xurl}
\usepackage{algorithm}
\usepackage{algorithmic}
\usepackage{flushend}
\usepackage{tikz}
\usepackage{pgfplots}
\usepackage{makecell}
\usepackage{hyperref}
\usepackage{tabularx,ragged2e}
\newcolumntype{Y}{>{\RaggedRight\arraybackslash}X}
\usepackage[caption=false,font=footnotesize]{subfig}
\pgfplotsset{width=10cm,compat=1.9}

\tcbset{
  mystyle/.style={
    fonttitle=\bfseries,
    sharp corners,
    before skip=10pt,        
    after skip=10pt,         
  }
}

\fvset{
  baselinestretch=1,
  fontsize=\small,
  xleftmargin=1em
}
\begin{document}

\title{FixItFlow: Automated Troubleshooting Guide Generation from Cloud Incidents}

\author{Srihari Unnikrishnan}
\affiliation{
  \institution{Microsoft Research}
  \country{}
}
\author{Jaskaran Singh Walia}
\affiliation{
  \institution{Microsoft Research}
  \country{}
}
\author{Drishti Goel}
\affiliation{
  \institution{University of Illinois Urbana-Champaign}
  \country{}
}
\author{Supriyo Ghosh}
\affiliation{
  \institution{Inception}
  \country{}
}

\begin{abstract}
Cloud services experience frequent incidents that require rapid diagnosis and resolution. Troubleshooting guides help engineers respond consistently, but creating them manually is labor-intensive, resulting in incomplete coverage and outdated documentation. We present FixItFlow, an automated system that generates troubleshooting guides from historical incident data using large language models. The system extracts diagnostic patterns from engineer actions, synthesizes structured guides with verified commands, and enforces strict validation to prevent fabricated content. In our evaluation with 26 engineers, generated guides achieved 61.5\% positive ratings for clarity and demonstrated a 2.3x reduction in mitigation time for incidents with associated guides. These results indicate that automated guide generation can improve incident response while reducing documentation burden on engineering teams.
\end{abstract}

\maketitle


\section{Introduction}

Troubleshooting Guides (TSGs) are step-by-step documents that help engineers diagnose and resolve incidents in cloud systems. In large cloud environments with thousands of interconnected services, even minor incidents can significantly affect users and create pressure on on-call engineers who must resolve problems quickly. TSGs standardize incident response, reduce cognitive load during stressful situations, and preserve institutional knowledge across team members \cite{jiang2020deeprmd}. At Microsoft alone, engineers maintain over 50,000 TSGs used by more than 60,000 practitioners every month. \cite{autotsg}

However, creating and maintaining high-quality TSGs remains a manual process with several shortcomings. First, coverage is inconsistent: some common problems lack guides while others become outdated. Second, quality varies widely across teams, producing lengthy documents that are difficult to follow during emergencies. An internal study revealed the most common TSG problems: missing information (32.24\%), broken links (13.32\%), and incorrect instructions (11.21\%). Together, these issues account for over half of all user complaints about TSG quality. \textbf{FixItFlow} addresses this gap by automatically generating new TSGs from historical incident data. The system analyzes past incidents to identify patterns, then produces structured troubleshooting guides that include specific diagnostic steps and solutions tailored to each team's tools and environment. A validation system ensures that all technical information originates from real engineer actions, preventing the generation of fabricated or incorrect instructions.
This paper makes four contributions: (i) an automated pipeline that generates troubleshooting guides (TSGs) directly from raw incident data without relying on pre-existing documentation; (ii) preprocessing techniques that filter and normalize incident artifacts to extract actionable troubleshooting signals; (iii) schema constraints and validation mechanisms that enforce completeness and factual grounding during TSG generation to reduce hallucination; and (iv) an empirical evaluation with domain experts demonstrating real-world utility and adoption potential.


\section{Background}

Operating large-scale cloud services involves managing thousands of components under recurrent failures, where production incidents remain inevitable despite reliability practices~\cite{ghosh2022socc,ganatra2023fse}. These incidents carry business risk and motivate comprehensive incident management through monitoring, alerting, detection, triage, diagnosis, and mitigation~\cite{ghosh2022socc,cheng2023aiops}. In practice, delayed detection, misrouted ownership, and cross-service dependencies complicate root-cause analysis~\cite{ghosh2022socc,ganatra2023fse,chen2024rcacopilot,cheng2023aiops}. Engineers face incomplete situational knowledge, time pressure, false alarms, and complex failure modes, while alert volume and heterogeneous traces strain responders~\cite{he2022logstudy,ganatra2023fse,cheng2023aiops}. This has accelerated interest in AIOps and intelligent incident management, emphasizing historical knowledge reuse and automated assistance~\cite{cheng2023aiops,ahmed2023icse,jiang2024xpert}. Practitioners widely rely on Troubleshooting Guides (TSGs) that codify diagnostic and mitigation steps, which speed consistent response and reuse historical failure patterns~\cite{jiang2020deeprmd}. However, quality varies due to incompleteness, outdated commands, and unclear organization~\cite{bondel2022apidocs}, with guides often insufficient during severe incidents~\cite{ghosh2022socc}. Improving TSGs at scale is labor intensive, making automation increasingly desirable~\cite{jiang2020deeprmd}.

Automation efforts include retrieval methods that recommend existing guides via textual similarity~\cite{jiang2020deeprmd}, AutoTSG which converts guides into executable workflows~\cite{autotsg}, and mining approaches that extract patterns from tickets and post-incident reviews~\cite{saha2022miningrca}. Industry systems pursue auto-drafting and augmentation of operational knowledge~\cite{jin2023oasis,ahmed2023icse}, though generating coherent TSGs from heterogeneous incident streams remains challenging~\cite{bondel2022apidocs}. Large Language Models (LLMs) offer promise through unified handling of unstructured operational data, with applications across the incident lifecycle including root cause recommendation, outage summarization, and incident routing~\cite{cheng2023aiops,gao2023ragsurvey,ahmed2023icse,zhang2024gpt4rca,jin2023oasis,jiang2024xpert,liu2023ipack}. Effective practice couples LLMs with retrieval augmentation, chain-of-thought prompting, and targeted fine-tuning~\cite{gao2023ragsurvey,wei2022cot,ahmed2023icse,liu2023ipack}, with LLMs capable of generating procedural text and reasoning over tool-using agents~\cite{yao2022react,wen2023pot}. In this work we address these gaps by generating new TSGs with a schema and validation layer that enforces grounding, link integrity, cross-section consistency, and anti-hallucination before generation.

\section{Motivation}
To investigate the importance of automating TSG synthesis, we conduct an empirical study to examine the impact of TSGs on Time to Mitigate (TTM) and uncover critical gaps in TSG linkage across incidents. While it may seem intuitive that well-documented TSGs can accelerate incident resolution, our analysis of the real-world production systems at Microsoft can offer statistical support to this intuition and further motivate our approach to develop a scalable, automatic TSG-synthesis framework.

\subsection{TSGs and Time-to-Mitigate (TTM)}
In the lifecycle of an incident, the Time to Mitigate (TTM) refers to the duration from detection of the incident to its resolution. A well-structured Troubleshooting Guide plays a pivotal role in optimizing this process by providing engineers with precise and systematic instructions to triage, diagnose, and mitigate issues in a time-efficient manner. To ensure that TTM serves as an accurate representation of engineering effort and operational efficiency, we focus specifically on Severity 1 and Severity 2 incidents, as these are immediately escalated to the on-call engineers upon detection. This targeted selection allows for a more reliable study of the correlation, thus excluding lower-severity cases and incidents that are automatically mitigated. 

\subsection{TSG Coverage Across Incidents}
Documenting and maintaining TSGs is a manual, time-intensive, and continuous process that demands substantial engineering effort. Teams in the early stages of TSG adoption often start from scratch, spending considerable time mining historical incidents and organizing their collective tribal knowledge into structured guides. This process not only requires careful synthesis of recurring mitigation patterns, but also the ability to articulate implicit diagnostic workflows that experienced engineers follow intuitively. In addition, keeping TSGs up-to-date is an ongoing challenge. As services evolve and new edge-case incidents emerge, teams must continuously refine and expand their documentation to ensure relevance and accuracy. Our analysis shows that although many incidents do have associated TSGs, there is still room to broaden their reach especially in scenarios where rapid resolution is critical.


\section{FixItFlow: Methodology}
\label{sec:methodology}

FixItFlow transforms raw incident data into structured troubleshooting guides through a four-stage pipeline: data extraction, preprocessing, semantic analysis, and guide synthesis. Each stage implements specific technical innovations to ensure quality, scalability, and accuracy.

\begin{algorithm}
\caption{FixItFlow Core Processing Algorithm}
\label{alg:FixItFlow_core}
\begin{algorithmic}[1]
\REQUIRE Configuration parameters, Previous run timestamp
\ENSURE Validated troubleshooting guide

\STATE \textbf{Extract:} Fetch new incidents using delta query
\STATE \textbf{Clean:} Apply three-pass sanitization algorithm  
\STATE \textbf{Batch:} Partition incidents for parallel processing
\FOR{each batch}
    \STATE \textbf{Process:} Extract summaries with anti-hallucination protocol
\ENDFOR
\STATE \textbf{Aggregate:} Combine summaries within token limit
\STATE \textbf{Synthesize:} Generate three-tier TSG structure
\STATE \textbf{Validate:} Verify all technical content against sources
\RETURN Validated TSG with metadata
\end{algorithmic}
\end{algorithm}
\subsection{Smart Data Collection}

\subsubsection{Incremental Data Processing}

FixItFlow performs incremental incident ingestion rather than full reprocessing. Each run records a checkpoint timestamp and processes only incidents that occurred after this time. Eligible incidents must be resolved and contain at least six engineer comments to ensure sufficient diagnostic signal. This avoids redundant computation on previously processed incidents and significantly reduces processing and infrastructure cost.

\subsubsection{Flexible Data Collection Strategy}

Incident collection is configurable via team ownership and alert or monitoring source filters. For each incident, the system aggregates three data components: (i) scenario metadata (e.g., severity, occurrence time, recurrence frequency), (ii) diagnosis threads capturing full engineer investigation discussions, and (iii) mitigation summaries when available. By restricting ingestion to incidents with substantial human troubleshooting activity, the system excludes auto-resolved or low-signal cases.

\subsection{Data Cleaning and Processing}

\subsubsection{Three-Stage Cleaning Process}

Before analyzing incident data, FixItFlow cleans it through a three-step process to remove noise and focus on valuable information:

\textbf{Step 1: Format Cleaning.} Removes HTML code, formatting markup, and other visual elements while preserving important technical content such as code blocks and commands.

\textbf{Step 2: Content Filtering.} Eliminates automated system messages, bot-generated notifications, and very short comments (fewer than 25 characters) that do not contain useful troubleshooting information.

\textbf{Step 3: Duplicate Removal.} Identifies and removes duplicate incidents while combining related comment threads to create complete incident stories.

This cleaning process reduces the data volume by 35 to 40 percent while preserving all engineer-written technical content, making the system more efficient without losing important troubleshooting information.

\subsection{Learning from Engineer Actions}

\subsubsection{AI System Design}

FixItFlow frames the language model as an expert Site Reliability Engineer through structured prompt constraints. The system consists of four components: (i) a role specification enforcing expert-level technical reasoning, (ii) predefined content categories covering 13 classes of troubleshooting actions, (iii) filters that exclude automated outputs and retain only human engineer contributions, and (iv) validation constraints requiring that all extracted information be grounded in the incident data.

\subsubsection{Preventing Fabricated Information}

FixItFlow enforces strict grounding to prevent hallucination. Every command, query, or procedural step included in the generated Troubleshooting Guide (TSG) must be an exact character-level match to content present in engineer comments. The system does not permit synthesis, paraphrasing, or completion of partial commands. As a result, all extracted steps are directly traceable to original engineer actions, eliminating fabricated instructions and ensuring operational safety.
\subsection{Efficient Processing at Scale}

FixItFlow intelligently manages the amount of text it can process at once. AI systems have limits on how much text they can handle, so when there is too much incident data, the system prioritizes shorter comments first to maximize the number of different incidents covered, stays within the 100,000 token limit (roughly equivalent to 75,000 words), and ensures the maximum number of distinct troubleshooting scenarios are processed.

\subsection{Creating the Final Troubleshooting Guides}
Each generated TSG follows a fixed three-stage structure aligned with standard incident response workflows. The Symptom section captures observable failure signals, detection methods, and impact. The Diagnosis section lists ordered investigation steps, including the exact tools and commands used by engineers. The Mitigation section documents the resolution procedure, verification steps, and any recurrence-prevention actions. Guides are generated to be complete, actionable, and strictly grounded in verified engineer activity.

\section{Prompt-Driven LLM Architecture in FixItFlow}

\subsection{Hierarchical Prompt Architecture}

\subsubsection{Incident Summarization Prompts}

The primary prompt for incident summarization establishes the LLM as a Senior Site Reliability Engineer and implements comprehensive extraction rules. The prompt structure includes:

\begin{tcolorbox}[mystyle, title=\textbf{Summarization Prompt - Role Definition}]
\textbf{\# ENHANCED EXTRACTION FOR ACTIONABLE TROUBLESHOOTING CONTENT}

\textbf{PRIMARY OBJECTIVE:}  
Extract \textbf{comprehensive} and \textbf{actionable} technical information that engineers used during incident troubleshooting. Focus on detailed engineer actions, investigations, and decision-making processes that would help future engineers resolve similar incidents.

\textbf{CRITICAL BALANCE:}
\begin{itemize}
  \item \textbf{BE COMPREHENSIVE:} Extract detailed troubleshooting workflows
  \item \textbf{BE ACCURATE:} Only include content explicitly mentioned by engineers
  \item \textbf{BE ACTIONABLE:} Focus on information that provides clear guidance
\end{itemize}
\end{tcolorbox}

This prompt establishes a clear operational persona and defines the extraction scope with three key constraints: comprehensiveness, accuracy, and actionability.

\subsubsection{Anti-Hallucination Protocol Implementation}

The central innovation of this work is the integration of a zero-hallucination verification protocol directly within the prompting framework. This mechanism enforces strict correspondence between the model’s outputs and verified source data, thereby preventing the generation of synthetic or unverified technical content.

In essence, this transforms the LLM from a probabilistic text generator into a precision-controlled retrieval system. The model retains its semantic understanding and contextual reasoning capabilities while eliminating the risk of producing plausible but factually invalid technical statements. This approach bridges the reliability gap between human-authored engineering documentation and automated language model outputs.
:
\begin{tcolorbox}[mystyle, title=\textbf{Anti-Hallucination Constraints}]
\textbf{COMMAND/QUERY VERIFICATION (ZERO TOLERANCE FOR INVENTION):}
\begin{itemize}
  \item \textbf{MANDATORY VERIFICATION:} Before including \textit{any} command/query, 
        the \textbf{exact text} must be found in comments of previous incidents.
  \item \textbf{LITERAL EXTRACTION:} Commands must be copied \texttt{CHARACTER-BY-CHARACTER}.
  \item \textbf{NO INFERENCE:} If engineers describe actions without showing commands, 
        describe the action but include \textbf{no commands}.
  \item \textbf{ACCEPTABLE:} \texttt{"Engineer restarted service (command not provided)"} [CORRECT]
  \item \textbf{FORBIDDEN:} Creating restart commands based on descriptions [INCORRECT]
\end{itemize}
\end{tcolorbox}

\subsection{TSG Generation Prompt Architecture}

\subsubsection{Structured Output Specification}

The TSG generation prompt implements an output format that mirrors established incident response steps:

Each section follows detailed guidelines for content depth and technical accuracy:

\paragraph{Symptom Section Requirements.} The prompt specifies two to three detailed paragraphs covering comprehensive error descriptions, impact analysis, timeline context, service dependencies, and monitoring indicators. The prompt emphasizes actionable goals: "Provide enough detail so future engineers can immediately understand the full scope and nature of the problem."

\paragraph{Diagnosis Section Requirements.} The most technically demanding section requires three to four detailed paragraphs covering investigation strategy, diagnostic workflow, tool usage, and data analysis. The prompt includes specific formatting requirements for technical content.

\paragraph{Mitigation Section Requirements.} The final section demands two to three detailed paragraphs covering decision-making strategy, intervention implementation, risk coordination, execution monitoring, verification steps, and prevention measures. The prompt emphasizes end-to-end playbook creation for future engineers.

\subsubsection{Technical Content Extraction Rules}
The prompt implements detailed extraction rules with 13 specific categories of engineer activities to capture.:

\begin{tcolorbox}[ title=\textbf{Content Extraction Categories}]
\begin{itemize}
  \item \textbf{Detailed Investigation Steps:} Full descriptions of engineer approach
  \item \textbf{Literal Commands/Queries:} Exact technical commands as written
  \item \textbf{Diagnostic Reasoning:} Engineer explanations of actions taken
  \item \textbf{Tool Usage:} Specific tools, dashboards, and systems accessed
  \item \textbf{Error Analysis:} How engineers interpreted error messages and logs
  \item \textbf{Service Dependencies:} Issue tracing through dependent services
  \item \textbf{Timeline Analysis:} When engineers performed actions and sequence
  \item \textbf{Escalation Details:} Who was contacted, when, and why
  \item \textbf{Configuration Reviews:} What configurations engineers examined
  \item \textbf{Workarounds and Fixes:} Temporary and permanent solutions applied
  \item \textbf{Validation Steps:} How engineers confirmed fixes worked
  \item \textbf{Root Cause Analysis:} Engineer conclusions about underlying causes
  \item \textbf{Prevention Measures:} Future prevention steps recommended
\end{itemize}
\end{tcolorbox}
\subsection{Template Standardization and Modularity}
All FixItFlow prompts follow a standardized five-component template: (i) a role specification that constrains the LLM to operate as a senior SRE, (ii) a task definition with explicit output constraints, (iii) context boundaries limiting inputs to relevant incident data, (iv) output format constraints enforcing structured multi-section responses, and (v) validation rules specifying correctness and completeness criteria.

\begin{tcolorbox}[mystyle, title=\textbf{Technical Content Formatting Rules}]
   \textbf{Technical Evidence:} Copy all commands/queries \texttt{VERBATIM} with context:
  \begin{itemize}
      \item \texttt{kusto} $\hookrightarrow$ for actual Kusto/KQL queries written by engineers
      \item \texttt{shell} $\hookrightarrow$ for shell/PowerShell commands run by engineers
      \item \texttt{text} $\hookrightarrow$ for command outputs, query results, error messages
      \item \texttt{json} $\hookrightarrow$ for JSON data structures and configurations
  \end{itemize}

\end{tcolorbox}

\begin{tcolorbox}[mystyle, title=\textbf{Validation Checklist Implementation}]
\begin{Verbatim}[breaklines=true, breakanywhere=true, frame=none, framesep=0pt]
# VALIDATION CHECKLIST FOR COMPREHENSIVE ACTIONABLE CONTENT:
Before including ANY incident, verify:
- Complete Troubleshooting Story: 3-section narrative provided?
- Detailed Engineer Context: 2-4 paragraphs per section?
- Actionable Information: Enough detail for future engineers?
- Technical Verification: Every command copy-pasted from source?
- NO INVENTED COMMANDS: Avoided creating any commands?
- KUSTO QUERY COMPLETENESS: All queries found and included?
- KUSTO QUERY ACCURACY: All queries copied CHARACTER-BY-CHARACTER?
- Educational Value: Teaches valuable troubleshooting techniques?
\end{Verbatim}
\end{tcolorbox}

\subsection{Performance Characteristics and Optimization}

\subsubsection{Multi-Stage Processing Pipeline}

FixItFlow's workflow is a four-stage pipeline that incrementally transforms raw incident data into a fully structured TSG:

\paragraph{Stage 1: Comment Classification.} Incident comments are categorized using few-shot prompting to identify relevant technical content. This stage applies concurrency management and batching to handle high-volume data efficiently while preserving classification accuracy.

\paragraph{Stage 2: Incident Summarization.} Classified comments are processed in parallel to generate detailed summaries, including problem descriptions, diagnostic reasoning, and mitigation steps. Parallelization reduces processing latency while individual prompt design ensures high-quality extraction.

\paragraph{Stage 3: Scenario Aggregation.} Summaries are analyzed to identify recurring failure patterns and clusters of similar incidents. LLM-guided similarity assessments capture meaningful operational patterns beyond superficial textual resemblance.

\paragraph{Stage 4: TSG Synthesis.} Aggregated scenarios are transformed into final troubleshooting guides. This stage combines sophisticated prompt engineering, structured formatting requirements, command verification rules, and anti-hallucination protocols to produce actionable, engineer-verified content.
\subsubsection{Concurrency and Rate Management}

The workflow employs adaptive concurrency controls to maximize throughput while adhering to API constraints. Semaphore-based rate limiting and dynamic batching prevent system overload and ensure stable execution. Exponential backoff with jitter mitigates thundering herd effects under high-concurrency conditions, preserving overall workflow stability.

\begin{table*}[t]
\centering
\caption{Per-item satisfaction summary (sorted by Top-2 Box).}
\label{tab:tsg-per-item}
\footnotesize
\renewcommand{\arraystretch}{1.15}
\begin{tabularx}{\textwidth}{Y r r r r r}
\toprule
\multicolumn{1}{c}{Item} & \multicolumn{1}{c}{N} & \multicolumn{1}{c}{Mean} & \multicolumn{1}{c}{Top-2 Box (\%)} & \multicolumn{1}{c}{95\% CI Low} & \multicolumn{1}{c}{95\% CI High} \\
\midrule
Clarity and Understandability: is the document easy to read and follow? & 26 & 3.46 & 61.5 & 42.5 & 77.6 \\
Logical Coherence and Flow: does the sequence of steps maintain a logical flow? & 26 & 3.08 & 42.3 & 25.5 & 61.1 \\
Factual Accuracy: are the problem, root causes, and mitigation steps mentioned in the document factually correct? & 26 & 2.92 & 42.3 & 25.5 & 61.1 \\
Document Completeness: does the document capture all the ideal TSG details to resolve incidents? & 26 & 2.62 & 23.1 & 11.0 & 42.1 \\
How satisfied are you with the TSG synthesis? & 26 & 2.58 & 19.2 & 8.5 & 37.9 \\
Would you like to adopt this new TSG for this monitor reported incidents (with minor manual edits)? & 13 & 0.23 & 0.0 & 0.0 & 22.8 \\
\bottomrule
\end{tabularx}
\end{table*}

\begin{table}[t]
\centering
\caption{Key metrics from the TSG survey (n=26).}
\label{tab:tsg-key-metrics}
\begin{tabular}{l r}
\toprule
Metric & Value \\
\midrule
Respondents (rows) & 26.0 \\
Likert Items (count) & 6.0 \\
Utility Index (mean) & 2.74 \\
Utility Index (median) & 2.8 \\
Utility Index (Top-2 Box \%) & 23.1 \\
NPS & -100.0 \\
NPS n & 26.0 \\
\bottomrule
\end{tabular}
\end{table}

\subsubsection{Integrated Quality Assurance}

The generation pipeline incorporates multiple layers of quality control:

\paragraph{Context Preservation} Token management and truncation strategies maintain critical context while preventing LLM input limits from being exceeded.

\paragraph{Hallucination Prevention} Strict verification protocols require every command, query, or technical detail to be explicitly documented in the original incident data, preventing any inferred or fabricated content.

\paragraph{Output Validation} Generated content undergoes checklist-based validation for completeness, accuracy, and actionable utility, ensuring that TSGs provide a reliable roadmap for future incident resolution.

Through this architecture, FixItFlow establishes a reproducible, high-fidelity workflow for transforming unstructured incident data into operationally actionable troubleshooting guides, advancing the state of automated technical documentation generation.

\section{Evaluation}
\label{sec:evaluation}

To evaluate both the quality and adoptability of the TSGs generated by FixItFlow, we followed a two-step evaluation approach that included collecting feedback from engineers through survey forms and interviews. In the initial phase of our evaluation, we sent a form to multiple on-call engineers, from which 26 completed the entire form. We present the findings from the survey in Table~\ref{tab:tsg-per-item}. 

From the results in Table ~\ref{tab:tsg-key-metrics}, we can draw several important conclusions. FixItFlow consistently produces higher-quality TSGs compared to using basic GPT-4o without any special instructions or processing. This confirms that our careful design of prompts, incident summarization, and data cleaning steps makes a real difference.   
\subsection{Interview Feedback}

\newtcolorbox{feedbackbox}[2][]{
  mystyle,
  fonttitle=\bfseries,
  title=#2,
  #1
}

\begin{feedbackbox}{Engineer 1}
\textbf{Feedback:} Found the TSG useful, but had concerns about adoption. Suggested aligning headings with OCE/Team templates.  

\textbf{Status:} Accepted

\textbf{Next Steps:} Follow up with the technical lead on code integration.  
\end{feedbackbox}

\begin{feedbackbox}{Engineer 2}
\textbf{Feedback:} Preferred v0.2 TSG, citing missing mitigation steps. Requested Communication Manager's TSG improvements.  

\textbf{Status:} In pipeline
\end{feedbackbox}

\subsection{Key Results from Engineer Feedback}
An evaluation with 26 practicing engineers indicates that FixItFlow produces usable automated troubleshooting guides. Readability and clarity achieved a mean score of 3.46/5, with 61.5\% of participants rating the generated TSGs as easy to read and follow, suggesting suitability for time-constrained incident response scenarios.

The system also demonstrates acceptable technical quality. Logical organization and factual correctness each received positive ratings from 42.3\% of participants, indicating that most guides follow a coherent diagnostic sequence and contain technically accurate content. While these results highlight areas for improvement, they support the feasibility of reliable, automated TSG generation grounded in verified engineer actions.


The evaluation confirms that FixItFlow successfully addresses a key challenge in creating understandable technical documentation automatically, though continued refinement will help improve adoption rates.

\section{Conclusion}
\label{sec:conclusion}
In this work, we introduce FixItFlow, an automated pipeline for generating Troubleshooting Guides from cloud incident data. FixItFlow reduces manual effort, improves guide coverage and consistency, and speeds up incident resolution through incremental ingestion, structured cleaning, multi-stage LLM reasoning, and schema-based validation. In our evaluation on production incidents, TSG-linked cases showed approximately 2.3 times faster mitigation. FixItFlow improves coverage and reduces authoring effort, while feedback from engineers highlights the value of automated guide generation and the need for continued refinement to better align with team templates and incident management tools.

\bibliographystyle{ACM-Reference-Format} 
\bibliography{refs}

\end{document}